# Improvement of AMPs Identification with Generative Adversarial Network and Ensemble Classification

Reyhaneh keshavarzpour and Eghbal G. Mansoori

School of Electrical and Computer Engineering,
Shiraz University, Shiraz, Iran

reyhaneh.keshavarz@hafez.shirazu.ac.ir , mansoori@shirazu.ac.ir

**Abstract**: Identification of antimicrobial peptides is an important and necessary issue in today's era. Antimicrobial peptides are essential alternatives to antibiotics for biomedical applications and numerous other practical uses. These oligopeptides are useful in drug design and cause innate immunity against microorganisms. Artificial intelligence algorithms have played a significant role in facilitating the identification of these peptides. This research is improved by improving the proposed method in the field of antimicrobial peptides prediction. The suggested method is improved by combining the best coding method from different perspectives. In the following, a deep neural network is used to balance the imbalanced combined datasets. The results of this research show that the proposed method have a significant improvement in the accuracy and efficiency of the prediction of antimicrobial peptides and are able to provide the best results compared to the existing methods. These development in the field of prediction and classification of antimicrobial peptides, basically in the fields of medicine and pharmaceutical industries, have high effectiveness and application.

**Keywords**: Biomedical data, antimicrobial peptide, encoding, ensemble classification, deep neural network

## 1. Introduction

"This serious threat is no longer a prediction for the future; it is happening right now in every region of the world and has the potential to affect anyone, of any age, in any country. Antibiotic resistance—when bacteria change so antibiotics no longer work in people who need them to treat infections—is now a major threat to public health," according to a WHO report published in April 2014 [1].

A new chapter in medical history was initiated by Fleming's discovery of penicillin, which paved the way for the creation of numerous antibiotic classes. Due to their dramatic reduction in the death rates linked with bacterial illnesses, antibiotics have transformed medicine [2]. They are a cornerstone of contemporary medicine, used to treat a wide range of bacterial infections, from simple infections to serious ones [3]. Antibiotic resistance, one of the biggest risks to world health that humans currently face, has emerged as a result of the overuse and misuse of antibiotics [4]. Bacteria can adapt and change over time to become resistant to the effects of antibiotics, a process known as antimicrobial resistance (AMR).

Globally, the emergence of resistance to conventional antibiotics has led to a notable upsurge in research focused on developing novel and non-traditional anti-infective drugs for the market. The scientific community has taken a keen interest in antimicrobial peptides (AMPs), as seen by the sharp rise in published studies [5].

AMPs are little bioactive proteins with a molecular weight of less than 10 KDa that are typically made up of 10–50 amino acids. The majority of AMPs have a net positive charge of 2–13 and are mainly made of arginine and lysine (with some histidine thrown in), which together constitute a particular cationic domain [6] [7]. A few AMPs have a negative charge; two such AMPs are daptomycin and chrombacin, an antibacterial neuroendocrine peptide that has 12 net negative charges [8].
Although the term "AMP" is still accurate in the case of activity against bacteria [9], the term "Host Defence Peptide" (HDP) [10] [11]is now more widely used to underline the diverse nature of these molecules and represent the breadth of biological processes they regulate.

Based on where they came from, natural AMPs can be categorized as bacteriophage/viral AMPs, bacterial AMPs, fungal AMPs, plant-derived AMPs, and animal-derived AMPs [12].

AMPs, which are a component of the immune system by nature, are particularly prevalent on exposed surfaces like the skin and mucosa [5]. AMPs are expressed at these locations to act as a defense mechanism and shield the body against microbial invaders. The defense mechanisms cover a variety of bacterial interactions, mostly because of the physicochemical characteristics of the AMP and the ensuing three-dimensional structure. That is, peptides with lengths ranging from 10 to 50 residues are primarily composed of positively charged and hydrophobic residues, which can form α-helices, β-sheets, or random coils [5]. Because of the "multi-hit mechanism," it is challenging for viruses to adapt to AMPs, which makes them effective even against extremely resistant ones.

Active peptides engage in two different interactions with pathogens: first, they damage the bacterial membrane; second, they translocate, or move deeper inside the cell [13]. Because bacterial and eukaryotic membranes differ from one another, AMPs interact with their appropriate targets in a very selective manner [14]. Important ions and metabolites are lost as a result of the membrane breakdown, which ultimately causes cell lysis and cell death [5]. Translocation—the subsequent move to an intracellular location—occurs without breaching the pathogen's membrane. AMPs accumulate in the cytoplasm of the cell and prevent the production of proteins and nucleic acids [15]. There have also been reports of antiparasitic, antitumor, and antiviral properties in addition to antimicrobial ones [16].

## 2. Related works

AMPs can target a wide variety of harmful microbes, including viruses, parasites, bacteria, and fungi. They are produced by almost all known living species. While a number of studies have suggested various machine learning techniques to identify peptides as AMPs, the majority fail to take into account the variety of AMP actions [17].

Chung et al. (2019) carefully looked at the sequence characteristics of AMPs that target mammals as well as Gram-positive and Gram-negative bacteria. These AMPs have a variety of functional activities, such as anti-parasitic, anti-viral, anti-cancer, and anti-fungal properties [18].

A novel framework is put forth to methodically define, classify, and identify AMPs and the functional actions they perform. The suggested method consists of two stages: the first is identifying the AMPs, and the second is characterizing their functional activities in further detail. To extract potentially useful traits that may be connected to the functional activities of the AMPs, sequential forward selection was used. These characteristics include solvent accessibility, polarity, charge, hydrophobicity, and the normalized van der Waals volume, all of which are crucial for differentiating AMPs from non-AMPs [18].

In another study [19] authers used of AMPs as an example to show how to employ a VAE for de novo peptide sequence design. Similar to other computational techniques, the VAE outlined has the potential to automate sequence design as long as a sizable database of AMP sequences is available. Because there are more AMP databases and entries in each database, AMPs are a practical proof of concept for evaluating VAEs' capacity to produce novel peptides. The VAE's encoder and decoder are based on LSTM RNNs with high-dimensional sequence representations, which means that instead of just copying sequence templates, they also pick up crucial characteristics from the training dataset [19]. This strategy either uses genome-wide scanning to breed AMPs [20].

Empir was created as a quick and precise AMP classification framework to aid in the genome-wide identification of AMPs. When used with whole-genome data, Empir offers a significantly greater classification accuracy than current techniques and is built for high throughput. It also interacts well with existing bioinformatics processes.
Empir is mostly implemented in R, with C++ being used for the primary feature calculation functions [20].

Additional techniques involve employing instruments to investigate the connections among their structures, dynamics, and functions, as well as the growing utilization of machine learning and molecular dynamics simulations. Resources, including AMP databases, AMP-related web servers, and widely-used methods, are compiled in this study with the goal of helping researchers in the field complement experimental studies with computational approaches [21].

To enhance PIPs detection performance, Khatun et al. (2020) suggested a computational model termed ProIn-Fuse in conjunction with a multiple feature representation [22]. ProIn-Fuse predicted proinflammatory peptides by combining a fusion-based ensemble with several machine learning techniques, including AdaBoost and Naïve Bayes. In this situation, their fusion-based model performed noticeably better than alternative algorithms [22].

Plisson et al. (2020) produced machine learning models and techniques for detecting outliers, which provide reliable forecasts for the identification of AMPs and the creation of new peptides with decreased hemolytic activity [23].

On the other hand, Timmons et al. (2020) used advancements in machine learning research to create a novel artificial neural network classifier for the prediction of hemolytic activity from a peptide's main sequence in the HAPPEN method. [24].

Singh et al. (2021) conducted a comparative study of multiple ensemble approaches, such as GB and Extra Trees, and basic classifiers, such as Linear Discriminant Analysis, to identify AMPs. They proved that the GB had the best performance [25].

Chen et al. (2021) provide iLearnPlus, the first machine learning platform with online and graphical interfaces for building machine learning pipelines for protein and nucleic acid sequence analysis and prediction. Without the need for programming, iLearnPlus offers an extensive collection of algorithms that automate sequence-based feature extraction and analysis, model building and deployment, predicted performance evaluation, statistical analysis, and data visualization [26].

iLearnPlus has many more machine learning algorithms than the existing solutions, including over twenty that cover multiple deep learning approaches and a large range of feature sets that encode information from the input sequences [26].

## 3. Materials and methods

### 3.1. Database

In this approach, utilized a portion of the extensive dataset that Spänig et al. [27]. This priceless collection contains a wide range of data from several biomedical fields. Among this diverse set of peptides, this paper concentrated on those that fit into the following categories:

1. Anti-cancer (acp_mlacp): Peptides that may have anti-cancer qualities are essential in the continuous fight against cancer.

2. Anti-inflammatory (aip_antiinfam): Peptides with anti-inflammatory properties that help control a range of inflammatory ailments.

3.Anti-microbial (amp_antibp2): Peptides with anti-microbial properties that show promise in the fight against infectious diseases.

4.Cell-penetrating (cpp_mlcpp): Peptides that can penetrate cells are essential for intracellular targeting and medication administration.

5. Hemolytic Peptides (hem_hemopi): These peptides are essential for research on blood cell interactions and membrane disruption.

6. Immunosuppressive (isp_il10pred): Peptides with immunosuppressive characteristics that are important for immune modulation studies.

7. Pro-inflammatory (pip_pipel): Peptides that have pro-inflammatory properties and help us comprehend how the body reacts to inflammation [27].

The current study profited from a wide range of peptide kinds, each with a distinct biomedical value, by utilizing this large dataset. Because of the wide range of peptide categories, authors were able to investigate and evaluate numerous facets of peptide function and their uses in diverse biomedical contexts.

Table 1 . Additional details regarding each dataset

| | Dataset | Size(-,+) |
|---|---|---|
| 1 | acp_mlacp | 585 (398 , 187) |
| 2 | aip_antiinfam | 2124 (1261, 863) |
| 3 | amp_antibp2 | 1993 (994, 999) |
| 4 | atb_antitbp | 492 (246 , 246) |
| 5 | avp_amppred | 1478 (739 , 739) |
| 6 | cpp_mlcpp | 1903 (1165, 738) |
| 7 | hem_hemopi | 1104 (582, 522) |
| 8 | isp_il10pred | 1242 (848 , 394) |
| 9 | nep_neuropipred | 1750 (875, 875) |
| 10 | pip_pipel | 3228 (2395, 833) |

## 3.2. Peptides encoding

Peptide encodings' progress and improvement made it possible for machine learning techniques to successfully anticipate AMPs. Encodings satisfy the needs of many machine learning algorithms by mapping amino acid sequences of varying lengths to numerical vectors of the same length [28].

Every one of these encodings is designed to accurately represent biological relationships, as well as inherent details found in the primary sequence and higher order confirmations. Not only have several encodings been proposed, but different ways to integrate existing ones have also been put forth, because an informative encoding is critical to prediction accuracy [28].

This workflow contains 5 types of all encoding techniques:

### 3.2.1. Sparse encoding or AA sparse encoding

First, a peptide sequence is described using a technique called binary encoding, or SE for short. In SE, every amino acid is portrayed as a one-hot vector consisting of 20 elements, with all points being set to 0 except for one.

### 3.2.2. Amino acid composition

Representing the amino acid sequence as its matching composition is one technique to improve the density of the resulting feature space. According to this method, the feature vector contains the percentage of a specific amino acid to the total length of the sequence at each point [28].

### 3.2.3. Pseudo amino acid composition(PseAAC)

Such a set of sequence correlation factors and 20 amino acid components make up the pseudo-amino acid composition. Thus, the pseudo amino acid composition has the same mathematical values as the normal amino acid composition, with the exception of the compositional difference. The concept of amino acid-like composition (PseAAC) was developed in order to take sequence order into account [29]. A $\lambda + 20$ dimension vector is produced by PseAAC by computing the correlation between several ranges within a pair of amino acids. The first 20 components represent the makeup of the 20 naturally occurring amino acids, and components $20 + 1$ through $20 + \lambda$ explain the association in terms of the sequence order level at play [29].

### 3.2.4. Physicochemical properties

The conversion of amino acids into particular physicochemical properties is one of the key encoding schemes used in the prediction of AMPs; these properties have been painstakingly elucidated through numerous tedious wet lab experiments, highlighting their critical role in AMP prediction methodologies. The amino acid index database (AAindex) has been established as a unified source for these descriptors [30].

### 3.2.5. Fourier transform

By extending beyond the temporal domain and revealing the frequency spectrum, FT's elegant utility is demonstrated in its ability to reveal complex temporal patterns within time series data [31].

FT techniques have also been utilized successfully to investigate evolutionary relationships among genomic sequences [32]. Using Fourier transform techniques, the majority of protein science research has concentrated on structural issues. The binary indicator sequence method for DNA sequences and the EIIP index and hydropathy index [33] [34]method for protein sequences are the most widely used techniques for converting biological sequences into numerical sequences for signal processing. Within the field of biomedicine, FT is essential for tasks like determining the locations of repeating coding and non-coding portions in DNA sequences and predicting the cellular localization of proteins [35]. Furthermore, FT has been cleverly applied in a study that seeks to identify peptides with antibacterial potential. In this creative endeavor, the transformational FT process was applied after the amino acid residues were skillfully converted into physical characteristics [35].

## 3.3. Assessment

carried out a comparison analysis utilizing four different classifiers, namely Logistic Regression (LR), Random Forest (RF), Decision Tree (DT), and Gaussian Naïve Bayes (GN), in order to thoroughly assess the efficacy of our encoding approach to categorize AMPs. benchmarked these findings against the most popular base classifier used in the well-known method of the reference paper [36].

Classifiers that are often employed have a high mistake rate. Even though these mistakes are unavoidable, they can be minimized by building the learning classifier correctly. The process of creating several basic classifiers from which a new classifier is generated that performs better than each component classifier is known as ensemble learning. The training set, representation, meta-parameters, and algorithm employed by these base classifiers could vary [37].

The stack classifier is an ensemble learning technique that uses the output of several classifiers as input to a meta-classifier for the final classification task. The entire training set is used to train each classifier model, and the outputs, or meta-features, of each model are used to fit the meta-classifier [38].

In assessing the classification outcomes, a more dependable statistical rate is the Matthews Correlation Coefficient (MCC), which yields a high score only when the number of positive and negative elements in the dataset is proportionate to the prediction of good results in each of the four categories of the confusion matrix (true positive, true negative, false positive, and false negative). Despite their popularity, accuracy, and F1 score might yield deceptive results in unbalanced data sets because they neglect to account for the ratio of positive to negative factors. The MCC criterion is simple and easy to understand.
Regardless of their percentage in the whole dataset, the classifier must correctly predict the majority of positive cases and negative cases in order to receive a high quality score. Rather, only when applied to balanced data sets do F1 score and precision yield trustworthy conclusions [39].

MCC equation denoted as:

$$MCC = \frac{TP \times TN - FP \times FN}{\sqrt{(TP + FP)(TP + FN)(TN + FP)(TN + FN)}} \quad (1)$$

## 4. Proposed Stream

According to the figure, in alignment with a clearly defined objective,  A two-part workflow have been executed.

This manner makes the best use of five types of coding methods, including: sparse encoding, amino acid composition, amino acid pseudocomposition, properties of chemical properties, and Fourier transform. These methods were selected from all the encoding methods implemented [36].

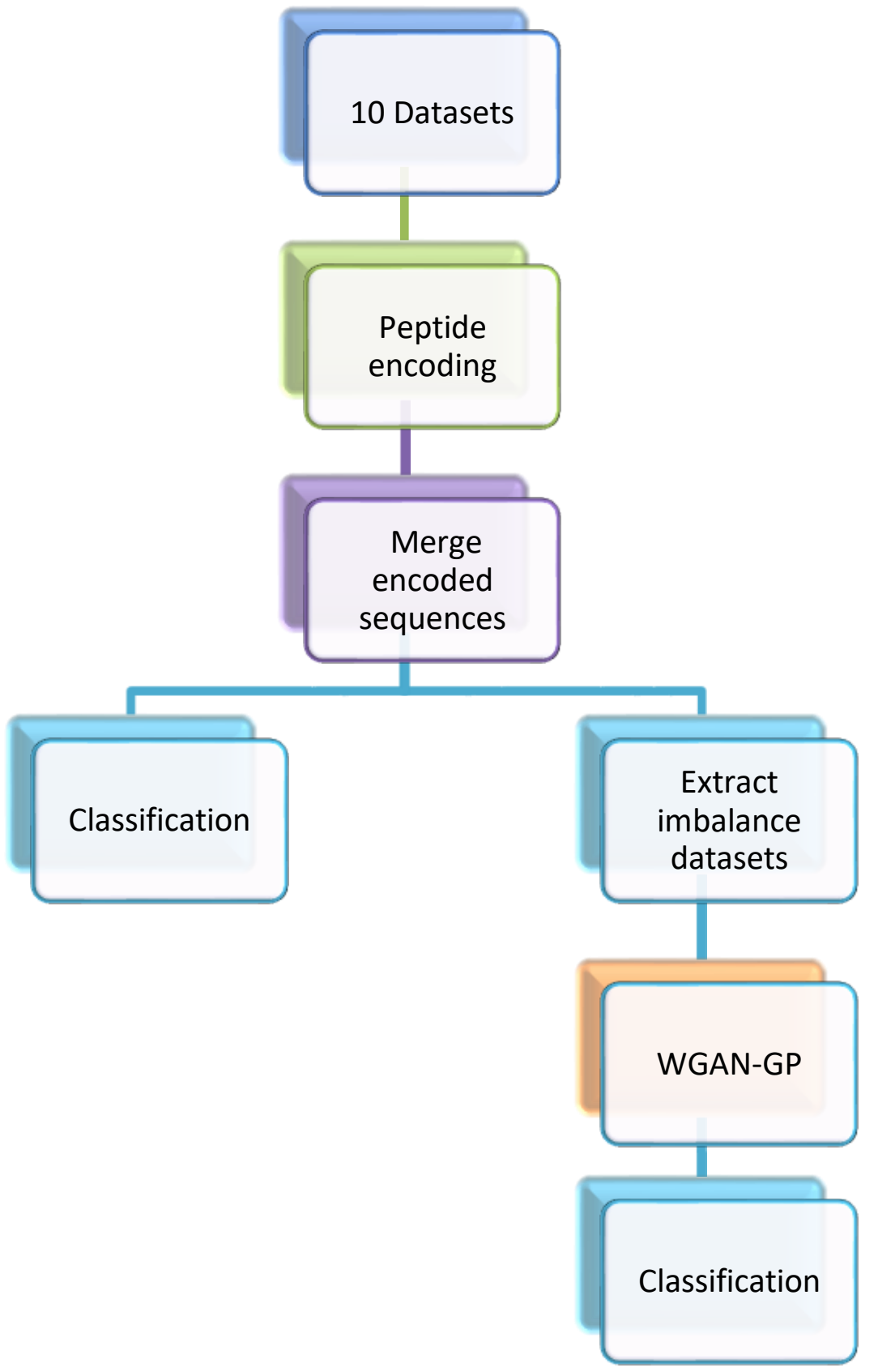

Figure 1 Show the two presented method in the form of two flowcharts

### 4.1. First Step Of Workflow

Consequently, in the methodological approach, the amino acid sequences are translated into numerical vectors. Following this translation, a consolidated dataset is created for each individual dataset by integrating the consensus of all encoding methods. This integrated approach ensures that the resultant numerical values comprehensively represent the original amino acid sequences, facilitating more accurate and reliable predictions in subsequent modeling stages.

To elaborate further, the translation of amino acid sequences into numerical vectors involves encoding each amino acid with specific numerical values that capture its biochemical properties and interactions. Various encoding schemes, such as one-hot encoding, physicochemical property-based encoding, and evolutionary information-based encoding, are employed to achieve this transformation. By integrating the consensus of all these encoding methods, we create a robust and comprehensive representation of the sequences.

The consolidated dataset thus generated encapsulates the essential features and nuances of the amino acid sequences, allowing for enhanced data quality. This integration process mitigates the biases and limitations inherent in individual encoding methods, thereby enriching the dataset's overall information content. Consequently, this comprehensive numerical representation serves as a solid foundation for subsequent machine learning and statistical modeling tasks.

In the subsequent modeling stages, the enriched dataset enables the application of advanced predictive algorithms with greater accuracy and reliability. The models trained on this dataset can leverage the detailed and holistic representation of the amino acid sequences, resulting in improved performance in tasks such as protein structure prediction, functional annotation, and interaction prediction. This methodological rigor ensures that the predictions derived from the models are both robust and generalizable, contributing significantly to advancements in computational biology and bioinformatics.

In the continuation of the presented approach, after concatenate encodings, 7 of the 10 datasets that are unbalanced (Table 1) have been balanced using the WGAN-GP deep learning model implemented by the PyTorch library. This network architecture can produce samples similar to real data. Imbalanced data sets are data in which the number of samples labeled 0 (samples with no antimicrobial activity) is greater than the number of samples labeled 1 (samples with antimicrobial activity).

In the WGAN-GP network structure, each linear layer is followed by a LeakyReLU activation function with a slope of 0.2. The activation function used after the output layer is the ReLU function. Additionally, both the generator and the critic contain three layers without batch normalization.

Five classification models, including logistic regression, random forest, naive Bayes, decision tree, and multilayer perceptron, were used to classify the data obtained from the concatenation of the balanced data by WGAN-GP. These models are developed to analyze and evaluate the best results based on the merged data and balanced data produced by WGAN-GP. Finally, the stacking approach was used to combine the outputs of the models from the two proposed approaches. In this context, the class probabilities predicted by the base classifiers are employed as features in the meta-classifier model. Logistic regression was used as a meta-model, leveraging a stack classifier from the scikit-learn library [41].

This comprehensive methodology aims to enhance the performance of classification by addressing the challenge of imbalanced data. By employing WGAN-GP [42] for data balancing. Finally, robust ensemble learning techniques like stacking are used that strive to improve the accuracy and reliability of the predictive models. The combination of these advanced techniques showcases a rigorous attempt to tackle the inherent biases and limitations associated with imbalanced datasets, ultimately leading to more effective and generalizable models for antimicrobial activity prediction.

This approach used a 5-fold Monte Carlo cross-validation (MCCV) to partition datasets [43]. The results are more robust, similar, and use all of the confusion matrix's components since the MCCV enhances generalization and reduces variance. Each fold consists of a single split, with 20% of the data used for testing and 80% for training. MCCV employs a sampling with replacement technique, which allows splits to contain identical data more than once, in contrast to k-fold cross-validation. Nevertheless, the train, validation, and split tests do not contain duplicate samples [43].

## 5. WGAN-GP

Due to the interaction between the cost function and the weight constraint, which causes vanishing or expanding gradients without adjusting the cutoff threshold, WGAN optimization is challenging [43]. The Lipschitz constraint is not appropriately enforced by weight cutoff, which also serves as a weight adjustment. It limits the model's capacity and makes it harder to model intricate functions. Any weight may take a while to reach its limit if the cutoff parameter is significant. Consequently, until optimization, it gets harder to train the critic. When there is a big number of layers, a little cut, or no batch normalization (like RNN), it can easily lead to vanishing gradients, and due to its simplicity and good performance, it will face a weight cut.

WGAN-GP enforces the Lipschitz requirement by using a gradient penalty rather than a weight cut. By including a gradient penalty element in the cost function (eq2), WGAN-GP enhances WGAN [42].

$$L = \mathbb{E}_{\boldsymbol{x}\sim P_g}[D(\boldsymbol{x})] - \mathbb{E}_{\boldsymbol{x}\sim P_g}[D(\boldsymbol{x})] + \lambda\, \mathbb{E}_{\boldsymbol{x}\sim P_x}[(\, \|\nabla_x\, D(\boldsymbol{x})\|_2 - 1)^2] \qquad (2)$$

## 6. Result and discussion

In this section, a comprehensive collection of results has been meticulously compiled and carefully analyzed. Detailed comparisons between the various findings are provided to highlight the significant trends, similarities, and differences observed. By examining these comparisons, readers can gain valuable insights into the underlying patterns and implications of the study.

Figure 2 compares the performance of various encoding methods (e.g., Concat, Fourier Transform, Physico-chemical, Pseudo-AAC, AA Composition, and Sparse Encoding) across multiple datasets. It highlights the impact of balancing the datasets using GANs on the performance of these encoding methods. The performance is measured using a specific metric (MCC), and the results are shown for both the original datasets (-G) and the GAN-balanced datasets (+G).In Table 2, the five most effective coding methods for amino acid sequences are

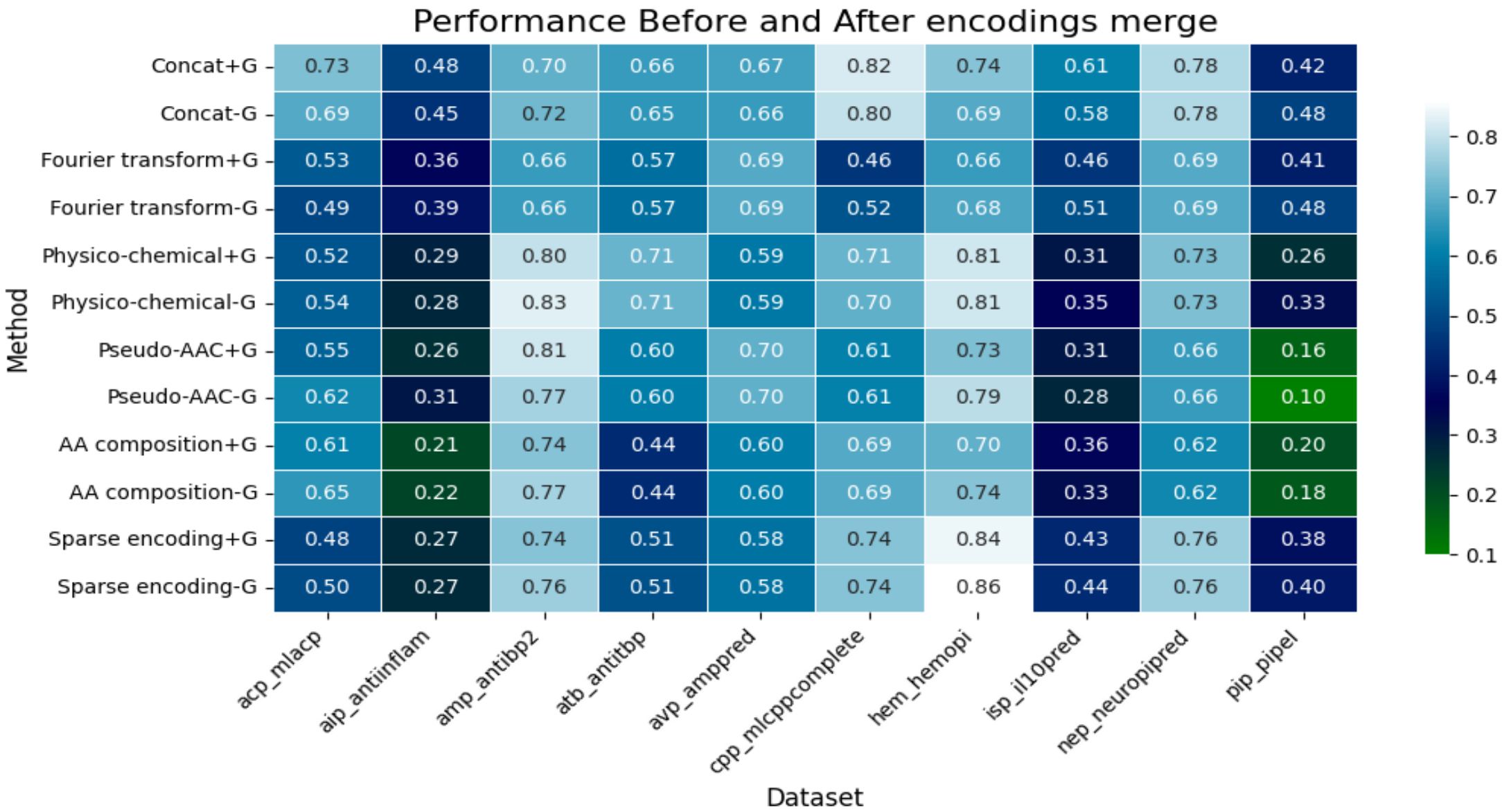

Figure 2 provides a comprehensive comparison of the performance of different encoding methods before and after balancing with GANs by using Random Forest classifier evaluated both before balancing (-G) and after balancing (+G) ( In the following tables, the data whose results are the same before and after balancing naturally have been balanced.).

Additionally, these coding methods are compared with their concatenated versions, considering both pre- and post-balancing scenarios (as shown in the last two columns).

Table 2 Results of MCC criterion for RF classification model on different datasets

| Datasets | Sparse encoding | | AA composition | | Pseudo-AAC | | Physico-chemical | | Fourier transform | | Concat | |
|---|---|---|---|---|---|---|---|---|---|---|---|---|
| | -G | +G | -G | +G | -G | +G | -G | +G | -G | +G | -G | +G |
| acp_mlacp | 0.50 | 0.48 | 0.65 | 0.61 | 0.62 | 0.55 | 0.54 | 0.52 | 0.49 | 0.53 | 0.69 | **0.73** |
| aip_antiinflam | 0.27 | 0.27 | 0.22 | 0.21 | 0.31 | 0.26 | 0.28 | 0.29 | 0.39 | 0.36 | 0.45 | **0.48** |
| amp_antibp2 | 0.76 | 0.74 | 0.77 | 0.74 | 0.77 | 0.81 | 0.83 | 0.80 | 0.66 | 0.66 | 0.72 | 0.70 |
| atb_antitbp | 0.51 | 0.51 | 0.44 | 0.44 | 0.60 | 0.60 | 0.71 | 0.71 | 0.57 | 0.57 | 0.65 | **0.66** |
| avp_amppred | 0.58 | 0.58 | 0.60 | 0.60 | 0.70 | 0.70 | 0.59 | 0.59 | 0.69 | 0.69 | 0.66 | **0.67** |
| cpp_mlcppcomplete | 0.74 | 0.74 | 0.69 | 0.69 | 0.61 | 0.61 | 0.70 | 0.71 | 0.52 | 0.46 | 0.80 | **0.82** |
| hem_hemopi | 0.86 | 0.84 | 0.74 | 0.70 | 0.79 | 0.73 | 0.81 | 0.81 | 0.68 | 0.66 | 0.69 | **0.74** |
| isp_il10pred | 0.44 | 0.43 | 0.33 | 0.36 | 0.28 | 0.31 | 0.35 | 0.31 | 0.51 | 0.46 | 0.58 | **0.61** |
| nep_neuropipred | 0.76 | 0.76 | 0.62 | 0.62 | 0.66 | 0.66 | 0.73 | 0.73 | 0.69 | 0.69 | 0.78 | 0.78 |
| pip_pipel | 0.40 | 0.38 | 0.18 | 0.20 | 0.10 | 0.16 | 0.33 | 0.26 | 0.48 | 0.41 | **0.48** | 0.42 |

Table 2 indicates that almost all data sets have seen a drop in MCC coding outcomes following balancing. However, following the addition of five coding techniques—sparse ecoding, amino acid composition, PseAAC, physicochemical properties, and Fourier transformation—it is noted that, along with the results obtained prior to the balancing of the additional data, there has also been a notable improvement in the results of GAN, which are connected to the results of the additional data.

By eliminating redundancy and highlighting key features, sparse encoding makes data representation more effective. Understanding the relative amounts of each amino acid in a sequence can be critical for comprehending the structures and functions of proteins. This is where amino acid composition comes into play. By taking into account both local and global sequence-order impacts, PseAAC adds new details to the sequence. The biological and physical features of the amino acids are better captured by including their physicochemical properties, which enhances the data representation even further. Conversely, the Fourier transform moves the data into the frequency domain, making it possible to find periodic patterns in the sequences.

When these methods are combined, the overall quality of the data is improved, which improves the performance of GAN models. The richer data representation greatly helps these models, which are intended to produce new data samples that closely mirror the original data. Consequently, the improved MCC coding results after balancing can be attributed to the GANs' ability to generate more dependable and accurate outputs.

For clarity, the results are visualized in Figure 3, which shows a grouped bar chart with performance metrics for each model and dataset. The chart reveals that GAN-based balancing generally improves model performance, particularly for datasets with severe class imbalance. Ensemble models showed the most significant improvements, while Decision Trees exhibited more modest gains. These findings highlight the effectiveness of GAN-based balancing in addressing class imbalance and improving model robustness.. The primary aim of this table is to demonstrate the impact of incorporating various coding methods on the Matthews correlation coefficient (MCC) of the Random Forest (RF) model.

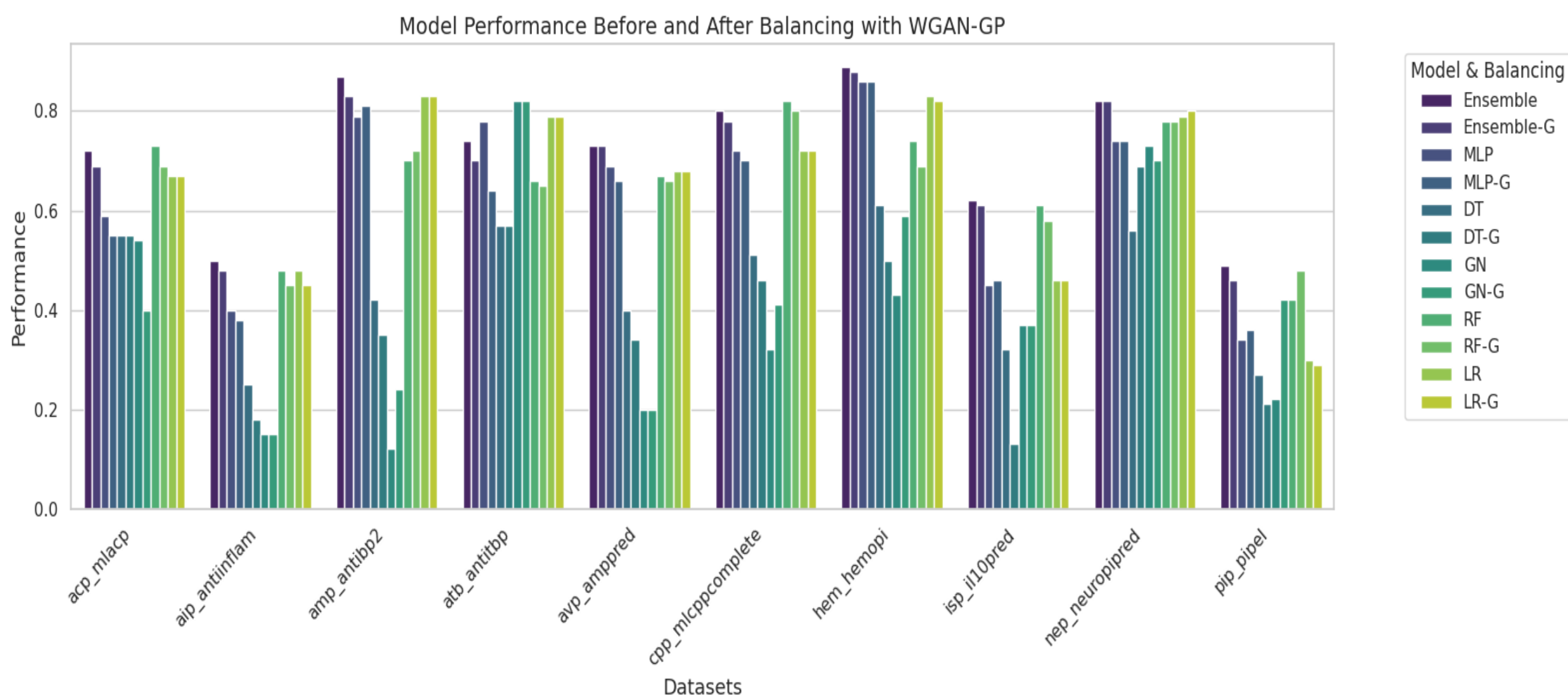

Figure3 The grouped bar chart visualizes the performance of various machine learning models across multiple datasets, comparing their performance before and after applying GAN-based data balancing

Table 3 The results of the MCC criterion for the model on different data sets

| Datasets | LR | | RF | | GN | | DT | | MLP | | Ensemble | |
|---|---|---|---|---|---|---|---|---|---|---|---|---|
| | -G | +G | -G | +G | -G | +G | -G | +G | -G | +G | -G | +G |
| acp_mlacp | 0.67 | 0.67 | 0.69 | 0.73 | 0.40 | 0.54 | 0.55 | 0.55 | 0.55 | 0.59 | 0.69 | 0.72 |
| aip_antiinflam | 0.45 | 0.48 | 0.45 | 0.48 | 0.15 | 0.15 | 0.18 | 0.25 | 0.38 | 0.40 | 0.48 | 0.50 |
| amp_antibp2 | 0.83 | 0.83 | 0.72 | 0.70 | 0.24 | 0.12 | 0.35 | 0.42 | 0.81 | 0.79 | 0.83 | 0.87 |
| atb_antitbp | 0.79 | 0.79 | 0.65 | 0.66 | 0.82 | 0.82 | 0.57 | 0.57 | 0.64 | 0.78 | 0.70 | 0.74 |
| avp_amppred | 0.68 | 0.68 | 0.66 | 0.67 | 0.20 | 0.20 | 0.34 | 0.40 | 0.66 | 0.69 | 0.73 | 0.73 |
| cpp_mlcppcomplete | 0.72 | 0.72 | 0.80 | 0.82 | 0.41 | 0.32 | 0.46 | 0.51 | 0.70 | 0.72 | 0.78 | 0.80 |
| hem_hemopi | 0.82 | 0.83 | 0.69 | 0.74 | 0.59 | 0.43 | 0.50 | 0.61 | 0.86 | 0.86 | 0.88 | 0.89 |
| isp_il10pred | 0.46 | 0.46 | 0.58 | 0.61 | 0.37 | 0.37 | 0.13 | 0.32 | 0.46 | 0.45 | 0.61 | 0.62 |
| nep_neuropipred | 0.80 | 0.79 | 0.78 | 0.78 | 0.70 | 0.73 | 0.69 | 0.56 | 0.74 | 0.74 | 0.82 | 0.82 |
| pip_pipel | 0.29 | 0.30 | 0.48 | 0.42 | 0.42 | 0.22 | 0.21 | 0.27 | 0.36 | 0.34 | 0.46 | 0.49 |

Table 3 shows the concatenation of the top five protein sequence coding techniques with the ensemble learning method before and after the balance, together with the MCC criterion results for all five classification models on different data sets. Following the comparison, the top result for each data set is displayed in the last two columns. The purpose of this table is to show how the ensemble learning of additive encodings influences MCC outcomes.
The results of the ensemble learning outperformed the performance of each utilized classification model for 70% of the data, as this table demonstrates.

The ensemble learning outcomes in this comparison, following further data balancing, are a significant improvement over the results of classification models alone.

Table 4 Results of MCC criteria for ensemble learning models and other models

| Dataset | Ensemblr learning before balance | Ensemblr learning after balance | Model 1 (2021) | Model 2 (2023) |
|---|---|---|---|---|
| acp_mlacp | 0.69 | 0.72 | 0.69 | 0.70 |
| aip_antiinflam | 0.48 | 0.50 | 0.45 | 0.49 |
| amp_antibp2 | 0.83 | 0.87 | 0.84 | 0.89 |
| atb_antitbp | 0.70 | 0.74 | 0.52 | 0.72 |
| avp_amppred | 0.73 | 0.73 | 0.8 | 0.79 |
| cpp_mlcppcomplete | 0.78 | 0.80 | 0.79 | 0.78 |
| hem_hemopi | 0.88 | 0.89 | 0.52 | 0.88 |
| isp_il10pred | 0.61 | 0.62 | 0.59 | 0.57 |
| nep_neuropipred | 0.82 | 0.82 | 0.67 | 0.81 |
| pip_pipel | 0.46 | 0.49 | 0.45 | 0.53 |

Table 4 shows a comparison between the results of the proposed method with competing methods [27] [36]. The obtained results indicate the positive effect of incorporating coding methods and balancing data using a deep neural network (GAN) in predicting antimicrobial peptides in most data sets compared to two competing methods.

## 7. Conclusion

Microbial infections that are resistant to chemically manufactured antibiotics have been created as a result of the overuse of antibiotics. Therefore, creating novel therapeutic strategies to treat infections is critically needed. This seemingly insurmountable obstacle indicates that the fast evolution of antibiotic-resistant bacteria—a problem that has turned into a global crisis—threatens to seriously undermine the substantial benefits of antibiotics. The use of artificial intelligence, particularly machine learning and deep learning algorithms, greatly simplifies the identification process and advances several antimicrobial peptides to the second and/or third phase of clinical trials, even though the efficiency of the process still needs to be verified in laboratory settings. Thus, the primary objective of this paper is to introduce a novel approach to enhance the identification of antimicrobial peptides for use in biomedicine and numerous other real-world contexts. By developing novel strategies for peptide discovery, this research aims to solve the problem of rising antibiotic resistance and open the road for advancement in the medical and multidisciplinary domains. Using a variety of amino acid sequence encoding techniques, we encoded each of the ten data sets examined in this research into a numerical vector. Once the encodings were added, to balance the uneven data that was divided into fivefold stratified cross-validation, we employed a deep neural network. With our approach, we have been able to improve the identification of antimicrobial peptides.

# References

[1] , WHO, "WHO's first global report on antibiotic resistance reveals serious, worldwide threat to public health," [Online]. Available: https://web.archive.org/web/20140502044726/http://www.who.int/en/.

[2] , B. W. W. A. M. I. K. M. M. S. R. M. H. .. &. B. Z. Aslam, "Antibiotic resistance: a rundown of a global crisis," *Infection and drug resistance,* pp. 1645-1658, 2018.

[3] , M. &. A. F. Gajdács, "Antibiotic resistance: from the bench to patients," *Antibiotics,* pp. 8(3), 129, 2019.

[4] , A. K. A. &. P. L. Baran, "Antibiotics and bacterial resistance—a short story of an endless arms race," *International Journal of Molecular Sciences,* pp. 24(6), 5777, 2023.

[5] , M. B. C. &. E. J. Mahlapuu, "Antimicrobial peptides as therapeutic agents: Opportunities and challenges," *Critical reviews in biotechnology,* pp. 40(7), 978-992, 2020.

[6] , A. L. P. R. T. B. d. S. Á. P. F. O. L. &. R. M. H. S. Lourenço, "Peptide Stapling Applied to Antimicrobial Peptides," *Antibiotics,* pp. 12(9), 1400, 2023.

[7] , C. D. H. J. A. H. R. E. &. S. G. Fjell, "Designing antimicrobial peptides: form follows function," *Nature reviews Drug discovery,* pp. 11(1), 37-51, 2012.

[8] , N. &. J. C. Chen, "Antimicrobial peptides: Structure, mechanism, and modification," *European Journal of Medicinal Chemistry,* pp. 255, 115377, 2023.

[9] , E. F. S. S. K. &. H. R. E. Haney, "Reassessing the host defense peptide landscape," *Frontiers in chemistry,* pp. 7, 435645, 2019.

[10] , A. &. H. R. E. Nijnik, "The roles of cathelicidin LL-37 in immune defences and novel clinical applications," *Current opinion in hematology,* pp. 16(1), 41-47, 2009.

[11] , D. S. S. K. P. O. &. Z. G. Takahashi, "Structural determinants of host defense peptides for antimicrobial activity and target cell selectivity," *Biochimie,* pp. 92(9), 1236-1241, 2010.

[12] , A. J. X. B. P. J. &. Z. Y. Bin Hafeez, "Antimicrobial peptides: an update on classifications and databases," *International journal of molecular sciences,* pp. 22(21), 11691, 2021.

[13] , J. O. C. G. F. F.-L. R. &. T. R. Cruz, "Antimicrobial peptides: promising compounds against pathogenic microorganisms," *Current medicinal chemistry,* pp. 21(20), 2299-2321, 2014.

[14] , E. Y. L. M. W. F. B. M. F. A. L. &. W. G. C. Lee, "What can machine learning do for antimicrobial peptides, and what can antimicrobial peptides do for machine learning?," *Interface focus,* pp. 7(6), 20160153, 2017.

[15] , F. V. N. A. P. D. L. D. S. S.-P. I. &. K. C. M. Guilhelmelli, "Antibiotic development challenges: the various mechanisms of action of antimicrobial peptides and of bacterial resistance," *Frontiers in microbiology,* pp. 4, 353, 2013.

[16] , Y. X. Q. Z. Q. H. Y. &. S. Z. Li, "Overview on the recent study of antimicrobial peptides: origins, functions, relative mechanisms and application," *Peptides,* pp. 37(2), 207-215, 2012.

[17] , J. H. C. Y. H. L. W. C. L. T. H. H. K. Y. &. L. T. Y. Jhong, "dbAMP: an integrated resource for exploring antimicrobial peptides with functional activities and physicochemical properties on transcriptome and proteome data," *Nucleic acids research,* pp. 47(D1), D285-D297, 2019.

[18] , C. R. K. T. R. W. L. C. L. T. Y. &. H. J. T. Chung, "Characterization and identification of antimicrobial peptides with different functional activities," *Briefings in bioinformatics,* pp. 21(3), 1098-1114, 2020.

[19] , S. N. &. W. S. A. Dean, "Variational autoencoder for generation of antimicrobial peptides," *ACS omega,* pp. 5(33), 20746-20754, 2020.

[20] , L. C. M. D. J. S. J. M. D. N. L. &. C. I. R. Fingerhut, "ampir: an R package for fast genome-wide prediction of antimicrobial peptides," *Bioinformatics,* pp. 36(21), 5262-5263, 2020.

[21] , P. G. R. L. M. D. N. L. J. F. S. J. Y. S. .. &. V. C. S. Aronica, "Computational methods and tools in antimicrobial peptide research," *Journal of Chemical Information and Modeling,* pp. 61(7), 3172-3196, 2021.

[22] , M. S. H. M. M. S. W. &. K. H. Khatun, "ProIn-Fuse: improved and robust prediction of proinflammatory peptides by fusing of multiple feature representations," *Journal of Computer-Aided Molecular Design,* pp. 34, 1229-1236, 2020.

[23] , F. R.-S. O. &. M.-H. C. Plisson, "Machine learning-guided discovery and design of non-hemolytic peptides," *Scientific reports,* pp. 10(1), 16581, 2020.

[24] , P. B. &. H. C. M. Timmons, "HAPPENN is a novel tool for hemolytic activity prediction for therapeutic peptides which employs neural networks," *Scientific reports,* pp. 10(1), 10869, 2020.

[25] , O. H. W. L. &. S. E. C. Y. Singh, "Co-AMPpred for in silico-aided predictions of antimicrobial peptides by integrating composition-based features," *BMC bioinformatics,* pp. 22, 1-21, 2021.

[26] , Z. Z. P. L. C. L. F. X. D. C. Y. Z. .. &. S. J. Chen, "iLearnPlus: a comprehensive and automated machine-learning platform for nucleic acid and protein sequence analysis, prediction and visualization," *Nucleic acids research,* pp. 49(10), e60-e60, 2021.

[27] , S. M. S. H. G. H. A. C. &. H. D. Spänig, "A large-scale comparative study on peptide encodings for biomedical classification," *NAR genomics and bioinformatics,* pp. 3(2), lqab039. https://doi.org/10.1093/nargab/lqab039, 2021.

[28] , S. &. H. D. Spänig, "Encodings and models for antimicrobial peptide classification for multi-resistant pathogens," *BioData mining,* pp. 12, 7. https://doi.org/10.1186/s13040-019-0196-x, 2019.

[29] , K. C. Chou, "Prediction of protein cellular attributes using pseudo-amino acid composition," *Proteins: Structure, Function, and Bioinformatics,* pp. 43(3), 246-255, 2001.

[30] , S. P. P. P. M. K. A. K. T. &. K. M. Kawashima, "AAindex: amino acid index database, progress report 2008," *Nucleic acids research,* pp. 36(suppl_1), D202-D205, 2007.

[31] , N. &. S. C. Strodthoff, "Detecting and interpreting myocardial infarction using fully convolutional neural networks," *Physiological measurement,* pp. 40(1), 015001, 2019.

[32] , M. D. P. D. &. S. M. A. Hendy, " A discrete Fourier analysis for evolutionary trees," *Proceedings of the National Academy of Sciences,* pp. 91(8), 3339-3343, 1994.

[33] , I. Cosic, "Macromolecular bioactivity: is it resonant interaction between macromolecules?-theory and applications," *IEEE Transactions on Biomedical Engineering,* pp. 41(12), 1101-1114, 1994.

[34] , J. &. D. R. F. Kyte, "A simple method for displaying the hydropathic character of a protein," *Journal of molecular biology,* pp. 157(1), 105-132, 1982.

[35] , V. K. N. M. B. Z. C. L. S. E. M. O. &. D. Y. Nagarajan, "A Fourier transformation based method to mine peptide space for antimicrobial activity," *BMC bioinformatics,* pp. (Vol. 7, pp. 1-8). BioMed Central, 2006, September.

[36] , S. M. A. &. H. D. Spänig, "Unsupervised encoding selection through ensemble pruning for biomedical classification," *BioData Mining ,* pp. 16, 10 .https://doi.org/10.1186/s13040-022-00317-7, 2023.

[37] , T. Acharya, "Advanced Ensemble Classifiers," 14 Jun 2019. [Online]. Available: https://towardsdatascience.com/advanced-ensemble-classifiers-8d7372e74e40.

[38] , M. Rutecki, "stacking classifier ensemble for great results," 2018. [Online]. Available: https://www.kaggle.com/code/marcinrutecki/stacking-classifier-ensemble-for-great-results.

[39] , D. &. J. G. Chicco, "The advantages of the Matthews correlation coefficient (MCC) over F1 score and accuracy in binary classification evaluation," *BMC genomics,* pp. 21, 1-13, 2020.

[40] , D. P. &. B. J. Kingma, "Adam: A method for stochastic optimization," *arXiv preprint arXiv:,* p. 1412.6980, 2014.

[41] , D. H. Wolpert, "Stacked generalization," *Neural networks,* pp. 5(2), 241-259, 1992.

[42] , I. A. F. A. M. D. V. &. C. A. C. Gulrajani, "Improved training of wasserstein gans," *Advances in neural information processing systems,* p. 30, 2017.

[43] , J. Hui, "GAN — Wasserstein GAN & WGAN-GP," 14 Jun 2018. [Online]. Available: https://jonathan-hui.medium.com/gan-wasserstein-gan-wgan-gp-6a1a2aa1b490.

[44] , S. A. Waksman, "What is an antibiotic or an antibiotic substance?," *Mycologia,* p. 565–569, 1947.

[45] , T. E. o. E. Britannica, "antibiotic," 10 June 2024. [Online]. Available: https://www.britannica.com/science/antibiotic.

[46] , M. A. P. K. Zilpah Sheikh, "Antibiotics: Everything You Should Know," 13 March 2024 . [Online]. Available: https://www.webmd.com/a-to-z-guides/what-are-antibiotics.

[47] , L. D., "Can better prescribing turn the tide of resistance?," *Nat Rev Microbiol,* p. 2:73–8, 2004.

[48] , N. E. D. A. &. B. N. M. Sabtu, "Antibiotic resistance: what, why, where, when and how?," *British medical bulletin,* p. 116(1), 2015.

[49] , E. ECDC, "The bacterial challenge–time to react a call to narrow the gap between multidrug-resistant bacteria in the eu and development of new antibacterial agents," *Solna: ECDC & EMEA Joint Press Release,* 2009.

[50] , U. D. o. H. a. H. Services, " Antibiotic resistance threats in the United States," 2019.

[51] , M. Akova, "Epidemiology of antimicrobial resistance in bloodstream infections," *Virulence,* pp. 7(3), 252-266, 2016.

[52] , M. F. R. L. &. R. R. Chellat, "Targeting antibiotic resistance," *Angewandte Chemie International Edition,* pp. 55(23), 6600-6626, 2016.

[53] , K. Q. M. H. Y. H. Huan Y., "Antimicrobial peptides:Classification, design, application and research progress in multiple fields," *Front. Microbiol,* p. 11, 2020.

[54] , L. J. P. I. K. M. K. S. Park C.B., " A novel antimicrobial peptide from the loach, Misgurnus anguillicaudatus," *FEBS Lett,* p. 411: 173–178., 1997.

[55] , C. Y. L. R. Lee I.H., "Styelins, broad-spectrum antimicrobial peptides from the solitary tunicate, Styela clava," *Comp.Biochem. Physiol. Part B: Biochem. Mol. Biol,* p. 118: 515–521, 1997.

[56] , Y. R. A. C. Reddy K.V.R., "Antimicrobial peptides:premises and promises," *Int. J. Antimicrob Agents, ,* p. 24: 536–547, 2004.

[57] , R.-S. L.A, "The role of amphibian antimicrobial peptides in protection of amphibians from pathogens linked to global amphibian declines," *Biophys Acta Biomembr ,* p. 1788: 1593–1599, 2009.

[58] , V. A., "Evolutionary plasticity of insect immunity," *J. Insect Physiol,* p. 59: 123–129, 2013.

[59] , D. M. M.-L. S. E.-M. M. Abdel-Latif H.M.R., "The nature and consequences of co-infections in tilapia: A review," *J. Fish Dis,* p. 43: 651–664, 2020.

[60] , K. Q. M. H. Y. H. Huan Y., "Antimicrobial peptides:Classification, design, application and research progress in multiple fields," *Front. Microbiol,* p. 11, 2020.

[61] , B. J. N. G. B. J. B. W.-J. A. Nawrot R., "Plant antimicrobial peptides," *folia Microbiol,* p. 59: 181–196, 2014.

[62] , K. J. S. G. K. P. Sharma P., "Plant derived antimicrobial peptides: Mechanism of target, isolation techniques sources and pharmaceutical applications," *J. Food Biochem.,* p. 46:e14348, 2022.

[63] , S. L. A. A. K. A. A. K. H. H. M. Nganso Y.O.D., "Identification of peptides in the leaves of Bauhinia rufescens Lam (Fabaceae) and evaluation of their antimicrobial activities against pathogens for aquaculture," *Science,* p. 8: 81–91, 2020.

[64] , H. J. H. W. Y. Y. C. M. G. H. T. Y. T. C. T. L. J. W. J. Hu S.Y., "Structure and function of antimicrobial peptide penaeidin-5 from the black tiger shrimp Penaeus monodon," *Aquaculture,* p. 260: 61–68, 2006.

[65] , Y. Y. T. M. A. A. Roch P., "NMR structure of mussel mytilin, and antiviral–antibacterial activities of derived synthetic peptides," *Dev. Comp. Immunol,* p. 32: 227–238, 2008.

[66] , S.-F. M. C. B. G. F. Valero Y., "Antimicrobial peptides from fish: beyond the fight against pathogens.," *Rev. Aquacult,* p. 12: 224–253, 2020.

[67] , turing, "Different Types of Cross-Validations in Machine Learning and Their Explanations," [Online]. Available: https://www.turing.com/kb/different-types-of-cross-validations-in-machine-learning-and-their-explanations.